\documentclass{tplp}
\usepackage{times}
\usepackage{helvet}
\usepackage{courier}
\usepackage{amsmath}
\usepackage{graphicx}
\usepackage{xspace}
\usepackage[defblank]{paralist}
\usepackage{latexsym}

\newcommand{\nesubset}{\rotatebox[origin=c]{45}{$\mathbf{\subset}$}}
\newcommand{\sesubset}{\rotatebox[origin=c]{-45}{$\mathbf{\subset}$}}

\title{Relative Expressiveness of Defeasible Logics}

\author[M.J. Maher]
{Michael J. Maher \\
School of Engineering and Information Technology \\
University of New South Wales, Canberra   \\
ACT 2600, Australia    \\
E-mail: m.maher@adfa.edu.au
}

\newcommand{\ignore}[1]{}
\newcommand{\finish}[1]{}

\newcommand{\DL}{{\bf DL}}

\newcommand{\ARROW}{\hookrightarrow}

\newcommand{\supp}{\sigma}   % support

\newcommand{\T}{{\cal T}}

\newcommand{\non}{{\sim}}

\newcounter{clause}
\def\theclause{$c$\arabic{clause}}

{\end{tabbing}}

{\end{tabbing}}

\newtheorem{theorem}{Theorem}

\newtheorem{definition}[theorem]{Definition}

\newtheorem{example}[theorem]{Example}
\begin{document}

\pagerange{\pageref{firstpage}--\pageref{lastpage}}
\volume{\textbf{10} (3):}
\jdate{March 2012}
\setcounter{page}{1}
\pubyear{2012}

\maketitle

\label{firstpage}

\begin{abstract}
We address the relative expressiveness of defeasible logics in the framework $\DL$.
Relative expressiveness is formulated as the ability to simulate the reasoning
of one logic within another logic.
We show that such simulations must be modular,
in the sense that they also work if applied only to part of a theory,
in order to achieve a useful notion of relative expressiveness.
We present simulations showing that logics in $\DL$ with and without
the capability of team defeat are equally expressive.
We also show that logics that handle ambiguity differently --
ambiguity blocking versus ambiguity propagating --
have distinct expressiveness, with neither able to simulate the other
under a different formulation of expressiveness.

\end{abstract}

\section{Introduction}

Defeasible reasoning concerns reasoning where a chain of reasoning
can be defeated (that is, not considered the basis of an inference)
by another chain of reasoning (or, perhaps, several chains of reasoning).
Defeasible logics are a class of non-monotonic logics designed to support defeasible reasoning.
Their rule-based approach is inspired by logic programming \cite{Nute_book}
and there is a close relationship between the logics and logic programming semantics \cite{MG99}.

Defeasible logics have some similarity to default logic \cite{DefaultLogic}.
An important difference is that default logic performs credulous reasoning,
whereas defeasible logics are sceptical.
In particular, an application of a default rule can proceed without reference to other default rules.
In contrast, in defeasible logics a rule can be applied only if all opposing rules are defeated.

The defeasible logics we address are distinguished by their choices on two orthogonal issues.
The first issue is one of \emph{team defeat}: when there are competing claims (on inferring $q$ or $\neg q$, say),
should a single claim for $q$ be required to overcome all competing claims in order to validate the inference,
or is it sufficient that every claim for $\neg q$ is overcome by some claim for $q$,
so that the claims for $q$, as a team, overcome all competing claims?
The second issue addresses \emph{ambiguity}, the situation where there is no resolution of the competing claims,
so that neither $q$ nor $\neg q$ can be derived.
Should ambiguity block, so that inferences relying on $q$ or $\neg q$ simply fail to apply,
or should the fact that there are claims for $q$ (say) that are not overcome by claims for $\neg q$
be allowed to influence later inferences, so that ambiguity propagates?
$\DL(\partial)$ and $\DL(\delta)$ are the ambiguity blocking and propagating logics, respectively, employing
team defeat, while $\DL(\partial^*)$ and $\DL(\delta^*)$ are the corresponding logics without team defeat.
The logics all fall within the $\DL$ framework \cite{flexf} of defeasible logics.

In this paper we investigate the notion of relative expressiveness with respect to the four logics named above.
Relative expressiveness can establish whether the different logics are substantively different, or simply provide the same capabilities in different formulations.
A logic that is less expressive than another does not require a dedicated implementation;
it can, in theory, be implemented via a translation to the more expressive logic\footnote{
In practice, however, since the logics addressed here have linear computational complexity \cite{Maher2001,TOCL10}, implementation by translation might not be as efficient as a direct implementation.
}.
On the other hand, a logic that is not less expressive than the other logics requires a separate implementation.
We explore alternative notions of relative expressiveness for defeasible logics
and make a considered choice of a formulation.

The main result is that logics with and without team defeat ({\it viz.} $\DL(\partial)$ and $\DL(\partial^*)$
and, separately, $\DL(\delta)$ and $\DL(\delta^*)$)
are equally expressive under this formulation.
This is surprising because team defeat appears a more sophisticated and powerful way
to adjudicate competing claims than requiring one claim to overwhelm all others.
It is also surprising because, in terms of relative inference strength,
$\DL(\partial)$ and $\DL(\partial^*)$ are incomparable,
and $\DL(\delta^*)$ is strictly weaker than $\DL(\delta)$.
A second result shows that the treatments of ambiguity are incomparable,
in terms of a different formulation of relative expressiveness.
This is also surprising when compared to relative inference strength.

The paper is structured as follows.
The next section provides an overview of defeasible logics.
It is followed by a discussion and formulation of the notion of simulation
that is central to our formulation of relative expressiveness.
The next two sections present a simulation of non-team defeat within a logic with team defeat
and, conversely, a simulation of team defeat within a logic without this capability.
In a section on ambiguity, we show that the two treatments of ambiguity are incomparable,
but in terms of a different notion of expressiveness.
Finally, we have a short discussion of the results and related work.

\section{Defeasible Logic}

In this section we can only present an outline of the defeasible logics we investigate.
Further details can be obtained from \cite{TOCL10} and the references therein.
We address propositional defeasible logics.
This might be restrictive in one sense, but a useful notion of relative expressiveness
should work on propositional logics as well as their first-order counterparts.

A \emph{defeasible theory} $D = (F, R, >)$ consists of a set of facts $F$, a finite set of rules $R$,
and a acyclic relation $>$ on $R$ called the \emph{superiority relation}.
This syntax is uniform for all the logics considered here.
Facts are individual literals
expressing indisputable truths.
Rules relate a set of literals (the body), via an arrow, to a literal (the head), and are one of three types:
a strict rule, with arrow $\rightarrow$;
a defeasible rule, with arrow $\Rightarrow$;
or
a defeater,  with arrow $\leadsto$.
Strict rules represent inferences that are unequivocally sound if based on definite knowledge;
defeasible rules represent inferences that are generally sound.
Inferences suggested by a defeasible rule may fail, due to the presence in the theory
of other rules.
Defeaters do not support inferences, but may impede inferences suggested by other rules.
The superiority relation provides a local priority on rules.
Strict or defeasible rules whose bodies are established defeasibly represent claims
for the head of the rule to be concluded.
The superiority relation contributes to the adjudication of these claims by an inference rule,
leading (possibly) to a conclusion.
For every theory $D$ there is a language $\Sigma$ containing all the literals addressed by $D$.  
We assume that $\Sigma$ is closed under negation.

Defeasible logics derive conclusions that are outside the syntax of the theories.
Conclusions may have the form 
$+d q$, which denotes that under the inference rule $d$ the literal $q$ can be concluded,
or
$-d q$, which denotes that the logic can establish that under the inference rule $d$ the literal $q$ cannot be concluded.
The syntactic element $d$ is called a tag.
In general, neither conclusion may be derivable:
$q$ cannot be concluded under $d$, but the logic is unable to establish that.
Tags $+\Delta$ and $-\Delta$ represent monotonic provability (and unprovability)
where inference is based on facts, strict rules, and modus ponens.
We assume these tags and their inference rules are present in every defeasible logic.
What distinguishes a logic is the inference rule for defeasible reasoning.
The four logics discussed in the Introduction correspond to four different pairs of inference rules,
labelled $\partial$, $\delta$, $\partial^*$, and $\delta^*$;
they produce conclusions of the form (respectively) $+\partial q$, $-\partial q$, $+\delta q$, $-\delta q$, etc. 
The inference rules $\delta$ and $\delta^*$ require auxiliary tags and inference rules,
denoted by $\supp$ and $\supp^*$, respectively.
For each of the four main defeasible tags $d$, the corresponding logic is denoted by $\DL(d)$.

At times we refer to a set of rules as a theory, 
implicitly choosing the set of facts and the superiority relation to be empty.
In general, every rule has a label with which to name it.
Labels are used in the superiority relation.
Where labels are not needed, they are omitted.
The \emph{size} of a theory is the total number of symbols used in expressing the theory.

The inference rules for $\DL(\partial)$ and $\DL(\partial^*)$ are presented 
below.
Given a defeasible theory $D = (F, R, >)$, for any set of conclusions $E$,
$\T_D(E)$ denotes the set of conclusions inferred from $E$ using $D$
and one application of an inference rule.
The inference rules are implicit in the definition of this function.
$\T_D$ is a monotonic function on the complete lattice of sets of conclusions ordered by containment.
The least fixedpoint of $\T_D$ is the set of all conclusions that can be drawn from $D$.
We follow standard notation in that $\T_D \uparrow 0 = \emptyset$ and $\T_D \uparrow (n+1) = \T_D(\T_D \uparrow n)$.

For every inference rule $+d$ there is a closely related inference rule $-d$
allowing to infer that some literals $q$ cannot be consequences of $D$ via $+d$.
The relationship between $+d$ and $-d$ is described as the Principle of Strong Negation \cite{flexf}.

Some notation in the inference rules requires explanation.
Given a literal $q$, its complement $\non q$ is defined as follows:
if $q$ is a proposition then $\non q$ is $\neg q$; if $q$ has form $\neg p$ then $\non q$ is $p$.
We say $q$ and $\non q$ (and the rules with these literal in the head)  \emph{oppose} each other.
$R_s$ ($R_{sd}$) denotes the set of strict rules (strict or defeasible rules) in $R$.
$R[q]$ ($R_s[q]$, etc) denotes the set of rules (respectively, strict rules) of $R$ with head $q$.
Given a rule $r$, $A(r)$ denotes the set of literals in the body of $r$.

{\small
\begin{minipage}[t]{.45\textwidth}
\begin{tabbing}
$+\Delta)$  $+\Delta q \in \T_D(E)$  iff  either  \\
\hspace{0.2in}  .1)  $q \in F$; or  \\
\hspace{0.2in}  .2)  $\exists r \in R_{s}[q]$  such that \\
\hspace{0.4in}      .1)  $\forall a \in A(r),  +\Delta a  \in  E$ \\
\end{tabbing}
\end{minipage}
\begin{minipage}[t]{.10\textwidth}
\hspace{1.0cm}
\end{minipage}
\begin{minipage}[t]{.45\textwidth}
\begin{tabbing}
$-\Delta)$  $-\Delta q \in \T_D(E)$  iff  \\
\hspace{0.2in}  .1)  $q \notin  F$, and \\
\hspace{0.2in}  .2)  $\forall r \in R_{s}[q]$  \\
\hspace{0.4in}      .1)  $\exists a \in A(r),  -\Delta a  \in  E$ \\
\end{tabbing}
\end{minipage}

%\smallskip
\noindent\begin{minipage}[t]{.45\textwidth}
\begin{tabbing}
$+\partial)$   $+\partial q \in \T_D(E)$  iff either \\
\hspace{0.2in}  .1)  $+\Delta q \in E$; or  \\
\hspace{0.2in}  .2)  The following three conditions all hold. \\
\hspace{0.4in}      .1)  $\exists r \in R_{sd}[q] \  \forall a \in A(r),  +\partial a \in E$,  and \\
\hspace{0.4in}      .2)  $-\Delta \non q \in E$,  and \\
\hspace{0.4in}      .3)  $\forall s \in R[\non q]$  either \\
\hspace{0.6in}          .1)  $\exists a \in A(s),  -\partial a \in E$;  or \\
\hspace{0.6in}          .2)  $\exists t \in R_{sd}[q]$  such that \\
\hspace{0.8in}              .1)  $\forall a \in A(t),  +\partial a \in E$,  and \\
\hspace{0.8in}              .2)  $t > s$.
\end{tabbing}
\end{minipage}
\begin{minipage}[t]{.45\textwidth}
\begin{tabbing}
$-\partial)$  $-\partial q \in \T_D(E)$  iff  \\
\hspace{0.2in}  .1)  $-\Delta q \in E$, and \\
\hspace{0.2in}  .2)  either \\
\hspace{0.4in}      .1)  $\forall r \in R_{sd}[q] \  \exists a \in A(r),  -\partial a \in E$; or \\
\hspace{0.4in}      .2)  $+\Delta \non q \in E$; or \\
\hspace{0.4in}      .3)  $\exists s \in R[\non q]$  such that \\
\hspace{0.6in}          .1)  $\forall a \in A(s),  +\partial a \in E$,  and \\
\hspace{0.6in}          .2)  $\forall t \in R_{sd}[q]$  either \\
\hspace{0.8in}              .1)  $\exists a \in A(t),  -\partial a \in E$;  or \\
\hspace{0.8in}              .2)  not$(t > s)$.\\
\end{tabbing}
\end{minipage}
\smallskip

\begin{minipage}[t]{.45\textwidth}
\begin{tabbing}
$+\partial^{*})$  $+\partial^* q \in \T_D(E)$  iff  either  \\
\hspace{0.2in}  .1)  $+\Delta q \in E$; or  \\
\hspace{0.2in}  .2)  $\exists r \in R_{sd}[q]$  such that \\
\hspace{0.4in}      .1)  $\forall a \in A(r),  +\partial^{*} a  \in  E$,  and\\
\hspace{0.4in}      .2)  $-\Delta  \non q  \in  E$,  and \\
\hspace{0.4in}      .3)  $\forall s  \in R[ \non q]$  either \\
\hspace{0.6in}          .1)  $\exists a \in A(s),  -\partial^{*}a  \in  E$;  or \\
\hspace{0.6in}          .2)  $r > s$.
\end{tabbing}
\end{minipage}
\begin{minipage}[t]{.07\textwidth}
\hspace{1.0cm}
\end{minipage}
\begin{minipage}[t]{.45\textwidth}
\begin{tabbing}
$-\partial^{*})$  $-\partial^* q \in \T_D(E)$  iff  \\
\hspace{0.2in}  .1)  $-\Delta q  \in  E$, and \\
\hspace{0.2in}  .2)  $\forall r \in R_{sd}[q]$  either \\
\hspace{0.4in}      .1)  $\exists a \in A(r),  -\partial^{*}a  \in  E$; or \\
\hspace{0.4in}      .2)  $+\Delta  \non q  \in  E$; or \\
\hspace{0.4in}      .3)  $\exists s  \in R[ \non q]$  such that \\
\hspace{0.6in}          .1)  $\forall a \in A(s),  +\partial^{*}a  \in  E$,  and \\
\hspace{0.6in}          .2)  not$(r > s)$. \\
\end{tabbing}
\end{minipage}
}
\begin{center}
\small Inference rules for $\Delta$, $\partial$, and $\partial^*$. 
\end{center}

In clause .1 of the inference rules for $+\partial$ and $+\partial^*$,
all literals derived from the monotonic part of the logic
are also considered defeasible consequences.
We can see that $+\partial^*$ relies on a single rule to overcome all competing rules.
At clause .2 a strict or defeasible rule $r$ must be found such that
all literals in the body have been established (.2.1) and,
every competing rule $s$ (.2.3) either fails to be established (.2.3.1)
or is inferior to $r$.
In comparison,
$+\partial$ relies on a team consisting of $r$ (.2.1) and all the rules $t$ (.2.3.2)
that are needed to overcome the competing rules $s$ (.2.3).

Thus $+\partial$ employs team defeat while $+\partial^*$ relies on a single rule overcoming all opposition.
For example, consider the following defeasible theory $D$ on whether animals are mammals \cite{TOCL01}.
\[
\begin{array}{lrll}
r_1:         & monotreme  &  \Rightarrow & mammal \\
r_2:         & hasFur  &  \Rightarrow &  mammal \\
r_3:         & laysEggs &  \Rightarrow & \neg mammal \\
r_4:         & hasBill &  \Rightarrow & \neg mammal \\
& r_1 > r_3 \\
& r_2 > r_4 \\
\end{array}
\]
For a platypus, we have the facts:
$monotreme$, 
$hasFur$,
$laysEggs$,
and 
$hasBill$.
The rules $r_3$ and $r_4$ for $\neg mammal$ are over-ruled by, respectively, $r_1$ and $r_2$.
Consequently, under inference with team defeat ($\partial$ and $\delta$),
we conclude $+\partial mammal$ and $+\delta mammal$.
Under inference without team defeat ($\partial^*$ and $\delta^*$),
there is no rule that overrules all the opposing rules.
Consequently we cannot make any positive conclusion;
we conclude $-\partial^* mammal$ and $-\partial^* \neg mammal$,
and similarly for $\delta^*$.

Both $\partial$ and $\partial^*$ are ambiguity blocking.
Consider the following theory $D$.
\[
\begin{array}{lrll}

r_1         &   &  \Rightarrow & p \\
r_2         &  &  \Rightarrow & \neg p \\
r_3         & &  \Rightarrow & q \\
r_4         & \neg p &  \Rightarrow & \neg q \\
\end{array}
\]
\noindent
$p$ and $\neg p$ are \emph{ambiguous}:
neither $r_1$ nor $r_2$ can overcome the other via the superiority relation.
Thus $-\partial \neg p$ is inferred.
Now, because the body of $r_4$ fails, there is no rule left to compete with $r_3$,
and so $+\partial q$ is inferred.
We also conclude $-\partial \neg q$; thus there is no ambiguity about $q$ and $\neg q$.
The same arguments apply for $\partial^*$.

On the other hand, $\delta$ and $\delta^*$ are ambiguity propagating.
$-\delta \neg p$ is inferred and consequently $-\delta \neg q$ is inferred.
However, ambiguity propagating logics like $\delta$ do not support a conclusion $+\delta q$.
There is a possibility that $\neg p$ holds, given that $r_2$ was not overcome via the superiority relation
but simply failed to overcome its competitor.
Hence there is a possibility that $\neg q$ holds.
And since $r_3$ cannot explicitly overcome $r_4$ via the superiority relation,
the conclusion $+\delta q$ is not justified and, in fact, $-\delta q$ is concluded.
This idea of ``possibly holding'' is called \emph{support};
it is expressed by an auxiliary tag $\supp$ and defined by a corresponding inference rule in $\DL(\delta)$
(and, similarly, the auxiliary tag $\sigma^*$ in $\DL(\delta)$).
In the theory $D$ above, among the conclusions are $+\supp p$, $+\supp \neg p$, $+\supp \neg q$, and $+\supp q$.
Since both $q$ and $\neg q$ possibly hold, they are ambiguous
and clearly the ambiguity has propagated.

A more detailed discussion of ambiguity and team defeat in the $\DL$ framework is given in \cite{TOCL10}.

Inference for $\delta$ and $\sigma$ (and $\delta^*$ and $\sigma^*$) is defined mutually recursively:

\smallskip
{\small
\smallskip
\noindent\begin{minipage}[t]{.45\textwidth}
\begin{tabbing}
$+\delta)$  If  $+\delta q \in \T_D(E)$  then either \\
\hspace{0.2in}  .1)  $+\Delta q \in E$; or  \\
\hspace{0.2in}  .2)  The following three conditions all hold. \\
\hspace{0.4in}      .1)  $\exists r \in R_{sd}[q] \  \forall a \in A(r),  +\delta a \in E$,  and \\
\hspace{0.4in}      .2)  $-\Delta \non q \in E$,  and \\
\hspace{0.4in}      .3)  $\forall s \in R[\non q]$  either \\
\hspace{0.6in}          .1)  $\exists a \in A(s),  -\sigma a \in E$;  or \\
\hspace{0.6in}          .2)  $\exists t \in R_{sd}[q]$  such that \\
\hspace{0.8in}              .1)  $\forall a \in A(t),  +\delta a \in E$,  and \\
\hspace{0.8in}              .2)  $t > s$.
\end{tabbing}
\end{minipage}
\begin{minipage}[t]{.45\textwidth}
\begin{tabbing}
$-\delta)$  If  $-\delta q \in \T_D(E)$  then \\
\hspace{0.2in}  .1)  $-\Delta q \in E$, and \\
\hspace{0.2in}  .2)  either \\
\hspace{0.4in}      .1)  $\forall r \in R_{sd}[q] \  \exists a \in A(r),  -\delta a \in E$; or \\
\hspace{0.4in}      .2)  $+\Delta \non q \in E$; or \\
\hspace{0.4in}      .3)  $\exists s \in R[\non q]$  such that \\
\hspace{0.6in}          .1)  $\forall a \in A(s),  +\sigma a \in E$,  and \\
\hspace{0.6in}          .2)  $\forall t \in R_{sd}[q]$  either \\
\hspace{0.8in}              .1)  $\exists a \in A(t),  -\delta a \in E$;  or \\
\hspace{0.8in}              .2)  not$(t > s)$.\\
\end{tabbing}
\end{minipage}

\begin{minipage}[t]{.45\textwidth}
\begin{tabbing}
$+\sigma)$  If  $+\sigma q \in \T_D(E)$  then either  \\
\hspace{0.2in}  .1)  $+\Delta q \in E$; or\\
\hspace{0.2in}  .2)  $\exists r \in R_{sd}[q]$  such that  \\
\hspace{0.4in}      .1)  $\forall a \in A(r),  +\sigma a \in E$,  and \\
\hspace{0.4in}      .2)  $\forall s \in R[\non q]$  either \\
\hspace{0.6in}          .1)  $\exists a \in A(s),  -\delta a \in E$;  or \\
\hspace{0.6in}          .2)  not$(s > r)$.
\end{tabbing}
\end{minipage}
\begin{minipage}[t]{.10\textwidth}
\hspace{1.0cm}
\end{minipage}
\begin{minipage}[t]{.45\textwidth}
\begin{tabbing}
$-\sigma)$  If  $-\sigma q \in \T_D(E)$  then \\
\hspace{0.2in}  .1)  $-\Delta q \in E$, and \\
\hspace{0.2in}  .2)  $\forall r \in R_{sd}[q]$  either \\
\hspace{0.4in}      .1)  $\exists a \in A(r),  -\sigma a \in E$;  or \\
\hspace{0.4in}      .2)  $\exists s \in R[\non q]$  such that \\
\hspace{0.6in}          .1)  $\forall a \in A(s),  +\delta a \in E$,  and \\
\hspace{0.6in}          .2)  $s > r$.\\
\end{tabbing}
\end{minipage}

\begin{minipage}[t]{.45\textwidth}
\begin{tabbing}
$+\delta^{*})$  If  $+\delta^{*}q \in \T_D(E)$  then either \\
\hspace{0.2in}  .1)  $+\Delta q  \in  E$; or  \\
\hspace{0.2in}  .2)  $\exists r  \in R_{sd}[q]$  such that \\
\hspace{0.4in}      .1)  $\forall a \in A(r),  +\delta^{*}a  \in  E$,  and \\
\hspace{0.4in}      .2)  $-\Delta  \non q  \in  E$,  and \\
\hspace{0.4in}      .3)  $\forall s  \in R[ \non q]$  either \\
\hspace{0.6in}          .1)  $\exists a \in A(s),  -\sigma ^{*}a  \in  E$;  or \\
\hspace{0.6in}          .2)  $r > s$.
\end{tabbing}
\end{minipage}
\begin{minipage}[t]{.10\textwidth}
\hspace{1.0cm}
\end{minipage}
\begin{minipage}[t]{.45\textwidth}
\begin{tabbing}
$-\delta^{*})$  If  $-\delta^{*}q \in \T_D(E)$  then \\
\hspace{0.2in}  .1)  $-\Delta q  \in  E$, and \\
\hspace{0.2in}  .2)  $\forall r \in R_{sd}[q]$  either \\
\hspace{0.4in}      .1)  $\exists a \in A(r),  -\delta^{*}a  \in  E$; or \\
\hspace{0.4in}      .2)  $+\Delta  \non q  \in  E$; or \\
\hspace{0.4in}      .3)  $\exists s  \in R[ \non q]$  such that \\
\hspace{0.6in}          .1)  $\forall a \in A(s),  +\sigma ^{*}a  \in  E$,  and \\
\hspace{0.6in}          .2)  not$(r > s)$.\\
\end{tabbing}
\end{minipage}

\begin{minipage}{.45\textwidth}
\begin{tabbing}
$+\sigma ^{*})$  If  $+\sigma ^{*}q \in \T_D(E)$  then either  \\
\hspace{0.2in}  .1)  $+\Delta q  \in  E$; or  \\
\hspace{0.2in}  .2)  $\exists r  \in R_{sd}[q]$  such that  \\
\hspace{0.4in}      .1)  $\forall a \in A(r),  +\sigma ^{*}a  \in  E$,  and \\
\hspace{0.4in}      .2)  $\forall s  \in R[ \non q]$  either \\
\hspace{0.6in}          .1)  $\exists a \in A(s),  -\delta^{*}a  \in  E$;  or \\
\hspace{0.6in}          .2)  not$(s > r)$.
\end{tabbing}
\end{minipage}
\begin{minipage}[t]{.10\textwidth}
\hspace{1.0cm}
\end{minipage}
\begin{minipage}{.45\textwidth}
\begin{tabbing}
$-\sigma ^{*})$  If  $-\sigma ^{*}q \in \T_D(E)$  then \\
\hspace{0.2in}  .1)  $-\Delta q  \in  E$, and \\
\hspace{0.2in}  .2)  $\forall r \in R_{sd}[q]$  either \\
\hspace{0.4in}      .1)  $\exists a \in A(r),  -\sigma ^{*}a  \in  E$;  or \\
\hspace{0.4in}      .2)  $\exists s  \in R[ \non q]$  such that \\
\hspace{0.6in}          .1)  $\forall a \in A(s),  +\delta^{*}a  \in  E$,  and \\
\hspace{0.6in}          .2)  $s > r$.
\end{tabbing}
\end{minipage}
}
\begin{center}
\small Inference rules for $\delta$, $\sigma$, $\delta^*$ and $\sigma^*$. 
\end{center}
\smallskip
\smallskip
\smallskip
\smallskip

There are surface similarities between defeasible logic and Reither's Default Logic \cite{DefaultLogic}, 
but there are also substantial differences.  
Default Logic employs a credulous semantics based on a model-theoretic view (the extensions), 
whereas defeasible logics take a proof-theoretic view.
Hence, from a defeasible theory $\Rightarrow p; \Rightarrow \neg p$ defeasible logics will not draw any conclusion\footnote{We refer only to positive conclusions, those using a tag $+d$.
}, whereas from the corresponding default theory
$\frac{: p}{p}, \frac{: \neg p}{\neg p}$ Default Logic will (separately) conclude each of $p$ and $\neg p$.
If the theory is extended by $p \Rightarrow q; \neg p \Rightarrow q$
then we find that defeasible logics will not draw any conclusion about $q$
while $q$ appears in both extensions of the corresponding default theory.
Thus, the formulation of scepticism through intersection of extensions is
different from the proof-theoretic formulation.
\cite{Horty} has a discussion of the two views, in the context of inheritance networks.

\section{Simulating Defeasible Logics}   \label{sect:sim}

A natural definition of relative expressiveness of logics
is to rely on the sets of conclusions that they are able to express.

\begin{definition}   \label{defn:sim}
The theory $D_1$ in logic $L_1$ is \emph{simulated} by $D_2$ in $L_2$
if
$D_1$ in $L_1$ and $D_2$ in $L_2$ have the same strict and defeasible conclusions,
modulo the tag that each logic uses.
We say $L_2$ is \emph{more (or equal) expressive} than $L_1$ if every theory $D_1$ in $L_1$
is simulated by some theory in $L_2$.
\end{definition}

That is, $D_2$ simulates $D_1$ when $D_1 \vdash \pm d_1 q$ iff $D_2 \vdash \pm d_2 q$,
where $\DL(d_1)$ is the logic of $D_1$ and $\DL(d_2)$ is the logic of $D_2$.
This appears to be quite a restrictive definition
since $D_2$ cannot use a larger language than $D_1$.
It also requires that the tag $d_1$ in $L_1$ is represented by the tag $d_2$ in $L_2$
rather than be expressed indirectly.
Nevertheless, it turns out to be a very coarse notion of relative expressiveness
that is unsuitable for separating the different logics in $DL$.

\begin{theorem}   \label{thm:simplesim}
For every pair of logics $L_1$ and $L_2$ in $DL$,
for every defeasible theory $D_1$ under the logic $L_1$
there is a defeasible theory $D_2$ under the logic $L_2$ that simulates $D_1$ under $L_1$.
\end{theorem}

The construction is straightforward.
Let $S$ be the set of conclusions from $D_1$ under $L_1 = \DL(d_1)$.
We construct $D_2 = (F, R, >)$ as follows, for every literal $q$:

If $+\Delta q \in S$ then add a fact $q$ to $F$.
If $+\Delta q \notin S$ and $-\Delta q \notin S$ then add $q \rightarrow q$ to $R$.
If $-\Delta q \in S$ then nothing is added to $D_2$.

If $+d_1 q \in S$ and $+\Delta q \in S$ then nothing is added.
If $+d_1 q \in S$ and $-\Delta q \in S$ then add $\Rightarrow q$ to $R$.
It is not possible for a logic infer $+d_1 q \in S$ and both $+\Delta q \notin S$ and $-\Delta q \notin S$,
so what to do in that case does not arise.
If $-d_1 q \in S$ then nothing is added.
If $+d_1 q \notin S$ and $-d_1 q \notin S$ then add $q \Rightarrow q$ to $R$.

Because of the simplicity of the theory $D_2$ that is constructed,
the conclusions of the theory are the same for all tags $d_2$.
Verification of the theorem requires checking that rules for $q$ and $\non q$ do
not interfere with each other.
For brevity, this part of the proof is omitted.

As a sidenote, observe that conclusions concerning support ($\supp, \supp^*$) have properties 
that cannot be simulated by the main tags.
In particular, the theory $\Rightarrow q; \Rightarrow \non q$ has as conclusions $+\supp q$ and $+\supp \non q$
(and similarly for $\supp^*$)
but none of the main tags can conclude both $+d q$ and $+d \non q$,
by the consistency property of these logics (Proposition 4 of \cite{TOCL10}).

In the previous theorem, the structure of the constructed theory $D_2$ is nothing like the theory $D_1$.
This freedom to choose $D_2$ without restriction is the reason why any theory in any logic
can be simulated by a theory in any other logic.
It is necessary to require that $D_2$ reflects some of the structure of $D_1$.
We do this indirectly, by requiring that the simulating theory be robust to certain changes.

We introduce the idea of an \emph{addition} $A$ to a theory $D$, denoted by $D + A$.
In general, $A$ is a defeasible theory: it may contain facts, rules and a superiority relation.
Let $D = (F, R, >)$ and $A = (F', R', >')$.
Then $D + A = (F  \cup F', R  \cup R', >  \cup >')$.

We would like to consider a theory $D'$ under logic $L'$
able to simulate $D$ under $L$ if the two theories have the same conclusions, modulo tags,
no matter what is added to both theories.
However, this is too strong a requirement.
For example, $p \rightarrow q$ is not simulated by $p \rightarrow t, t \rightarrow q$ under this definition
because the addition of the fact $t$ produces different behaviours.
We make two adjustments: we restrict additions to a class of theories,
and we allow $D'$ to ``hide'' internal symbols from interference by $A$.
We permit both $D'$ and $A$ to use symbols that are not used in $D$,
but we require that the symbols in $D'$ but not in $D$ are not available to $A$.
Thus we have the following \emph{language separation} condition:
$\Sigma(A) \cap \Sigma(D') \subseteq \Sigma(D)$.

\begin{definition}  \label{defn:simC}
Let $\Sigma$ be the language of the defeasible theory $D_1$.
Let $C$ be a class of defeasible theories $A$ such that $\Sigma(A) \cap \Sigma(D_2) \subseteq \Sigma(D_1)$.

We say $D_1$ in logic $L_1$ is \emph{simulated} by $D_2$ in $L_2$ with respect to a class $C$
if, for every addition $A$ in $C$,
$D_1 + A$ and $D_2 + A$ have the same conclusions in $\Sigma$, modulo tags.

We say a logic $L_1$ can be simulated by a logic $L_2$ with respect to a class $C$
if every theory in $L_1$ can be simulated by some theory in $L_2$ with respect to $C$.
\end{definition}

The use of a class $C$ gives us flexibility in expressing the degree of simulation
by varying the class, not only among those defined above, but many others.
For example, when $C$ consists only of the the empty theory $(\emptyset, \emptyset, \emptyset)$
the notion of simulation is notionally weaker than that of Definition \ref{defn:sim}
(weaker because it allows the simulation to use a larger language).
Larger classes of additions represent notionally stronger forms of simulation.

Consider addition limited to a set of facts, that is $A = (F, \emptyset, \emptyset)$.
Allowing arbitrary addition of facts corresponds to treating each theory $D_1$ under logic $L_1$
as defining a non-monotonic inference relation from facts to consequences.
This is similar to Dix's treatment of logic programs in \cite{Dix1} where
a logic program is viewed as defining a non-monotonic inference relation from the input atoms to the output atoms.
It also reflects a common practice of keeping the rules static while facts vary.
Simulation then requires that any inference relation expressed by $D_1$ under $L_1$ 
can be expressed by some $D_2$ under $L_2$.

However, it is not clear that the addition of facts is sufficiently discriminating.
For example,
we can attempt to extend the construction in Theorem \ref{thm:simplesim} by adding conditions
to the bodies of rules in that construction.
For a given defeasible theory $D$, we define $T(D)$ in several parts, as follows.
Define $NOT$ to be the set of all rules
\[
\begin{array}{lrl}
 & q & \Rightarrow \neg not\_q \\
 &    &  \Rightarrow not\_q
\end{array}
\]
for $q \in \Sigma$.
For any set $A \subseteq \Sigma$, we define $\hat{A}$ to be the conjunction of literals in the set
$A \cup \{ not\_a ~|~  D+A \vdash -d a\}$.  
$\hat{A}_{-q}$ denotes this conjunction with the omission of the literal $q$.
We also use $A$ as a conjunction of literals.

\noindent
We define $T(D, A)$ to contain:

\begin{tabular}[c]{rl}
$A \rightarrow q$ & if $D+A \vdash +\Delta q$ and $q \notin A$ \\

$A, q \rightarrow q$ & if $D+A \not\,\vdash +\Delta q$ and $D+A \not\,\vdash -\Delta q$ \\

$\hat{A}_{-q} \Rightarrow q$ & if $D+A \vdash +d q$ and $q \notin A$ \\

$\hat{A}_{-q}, q \Rightarrow q$ & if $D+A \not\,\vdash +d q$ and $D+A \not\,\vdash -d q$ \\
\end{tabular}

$T(D, A)$ describes the behaviour of $D + A$ in a way similar to the transformation in Theorem \ref{thm:simplesim},
but prefixes defeasible rules with $\hat{A}_{-q}$ to ensure that they are only applicable
when $A$ is the addition, or is a consequence of the addition.
For the strict rules, only a prefix $A$ is necessary, since any greater addition will also 
allow the inference of $+\Delta q$ (this is a reflection of the monotonicity of strict inference).

\noindent
We define $T(D)$ to consist of the facts from $D$ and the rules in $NOT \cup \bigcup_{A \subseteq \Sigma} T(D, A)$.
The superiority relation for $T(D)$ is empty.
In general, the size of $T(D)$ is exponential in the size of $D$.

While it is not proved that $T(D)$ simulates $D$,
the possibility of such a construction
prompts us to require that the simulating theory is limited to be of polynomial size,
and that the computation of the simulating theory can be achieved in polynomial time.
A \emph{polynomial simulation} of $L_1$ by $L_2$ is a mapping $T$ from each theory $D_1$
to a theory $D_2$ and a polynomial function $p(x)$ such that
$size(D_2) \leq p(size(D_1))$,
$t_{D_1 D_2} \leq p(size(D_1))$ where $t_{D_1 D_2}$ is the time to compute $D_2$ from $D_1$,
and
$D_2$ under $L_2$ simulates $D_1$ under $L_1$.
This leads us to a definition of relative expressiveness.

\begin{definition}   \label{def:expr}
A logic $L_1$ is \emph{more expressive} than a logic $L_2$ iff
there is a polynomial simulation of $L_2$ by $L_1$
with respect to the addition of facts.
\end{definition}

An alternative definition might use addition of rules, with or without the restriction to polynomial simulations.
The suitability of the current definition, and alternatives,
will depend on the results that can be obtained:
a notion of expressiveness that is so strict that no logic is more expressive than another,
or so lenient that all the logics have equivalent expressiveness,
has no practical use.
The following sections establish results showing that Definition \ref{def:expr} is not too strict.

\section{Team Defeat Simulates Non-Team Defeat}

We now show that every theory over a logic that does not employ team defeat
can be simulated by a theory over the corresponding logic that does employ team defeat.
Any defeasible theory $D$ is transformed into a new theory.
The new theory employs new propositions $h(r)$ for each rule $r$ in $D$,
and employs labels $p(r)$ for each rule $r$ in $D$
and labels $n(r, r')$ for each ordered pair of rules with opposing heads.

Let $D = (F, R, >)$ be a defeasible theory with language $\Sigma$.
We define the transformation $T$ of $D$ to $T(D) = (F', R', >')$ as follows:

\begin{enumerate}
\item
The facts of $T(D)$ are the facts of $D$.
That is, $F' = F$.
\item
For each rule $r = B \ARROW_r q$ in $R$,  $R'$ contains
\[
\begin{array}{lrll}
p(r): & B & \ARROW_r & h(r) \\
s(r):  & h(r) &  \rightarrow & q
\end{array}
\]

\noindent
and, further, for each rule $r' = B' \ARROW_{r'} \non q$ for $\non q$ in $R$, $R'$ contains

\[
\begin{array}{lrll}
n(r, r'):  & B' & \ARROW_{r'} & \neg h(r)
\end{array}
\]

\item
For every $r > r'$ in $D$, where $r$ and $r'$ are rules for opposite literals,
$T(D)$ contains $p(r) >' n(r, r')$ and $n(r', r) >' p(r')$.

\end{enumerate}

In this transformation, for each literal $q$, and for each rule $r$ for $q$,
we essentially create a copy of $r$ opposed to a copy of all rules for $\non q$
(the rules labelled $p(r)$ and $n(r, r')$ respectively).
$q$ is derived if a copy of some rule $r$ for $q$ is able to overcome the opposed rules
(that is, $q$ is derived without team defeat).

\begin{example}
To see the operation of this transformation, consider the following theory $D$:
\begin{tabbing}
1234123412341234\=123412341234\=1234\=1234\kill

\>$r_1: \  \ \Rightarrow p$
\>\> $r_3: \  \ \Rightarrow \neg p$ \\

\>$r_2: \ \ \Rightarrow p$
\>\> $r_4: \ \ \Rightarrow \neg p$ \\

\>$r_1 > r_3$ \>\>  $r_2 > r_4$

\end{tabbing}
In $\DL(\partial^*)$ from $D$ we conclude $-\partial^* p$ and $-\partial^* \neg p$, whereas
in $\DL(\partial)$ from $D$ we conclude $+\partial^* p$ and $-\partial^* \neg p$.
$T(D)$ contains the following rules and superiority relation.
\begin{tabbing}
12341234\=1234123412341234\=12341234123412341234\=12341234\kill

\>$p(r_1): \hspace{44pt}  \Rightarrow h(r_1)$
\>\> $p(r_3): \hspace{44pt} \Rightarrow h(r_3)$ \\

\>$s(r_1):\hspace{15pt} h(r_1) \rightarrow p$
\>\>$s(r_3):\hspace{15pt} h(r_3) \rightarrow \neg p$ \\

\>$n(r_1, r_3): \hspace{30pt} \Rightarrow \neg h(r_1)$
\>\>$n(r_3, r_1): \hspace{30pt} \Rightarrow \neg h(r_3)$ \\
\>$n(r_1, r_4): \hspace{30pt} \Rightarrow \neg h(r_1)$
\>\>$n(r_3, r_2): \hspace{30pt} \Rightarrow \neg h(r_3)$ \\

\ \\
\>$p(r_2): \hspace{44pt} \Rightarrow h(r_2)$
\>\> $p(r_4): \hspace{44pt} \Rightarrow h(r_4)$ \\
\>$s(r_2):\hspace{15pt} h(r_2) \rightarrow p$ 
\>\>$s(r_4):\hspace{15pt} h(r_4) \rightarrow \neg p$ \\

\>$n(r_2, r_3): \hspace{30pt} \Rightarrow \neg h(r_2)$
\>\>$n(r_4, r_1): \hspace{30pt} \Rightarrow \neg h(r_4)$ \\
\>$n(r_2, r_4): \hspace{30pt} \Rightarrow \neg h(r_2)$
\>\>$n(r_4, r_2): \hspace{30pt} \Rightarrow \neg h(r_4)$ \\
\ \\
\>$p(r_1) > n(r_1, r_3)$
\>\>$p(r_2) > n(r_2, r_4)$ \\
\>$n(r_3, r_1) > p(r_3)$
\>\>$n(r_4, r_2) > p(r_4)$ \\
\end{tabbing}
For each rule in $D$ there are four rules in $T(D)$.
Now 
$T(D) \vdash -\partial p$ and $T(D) \vdash -\partial \neg p$,
reflecting non-team defeat behaviour of $D$ within $\DL(\partial)$.
\end{example}

In general, the size of $T(D)$ is quadratic in the size of $D$.
Thus it remains to establish that $T(D)$ simulates $D$ with respect to addition of facts.

\begin{theorem}   \label{thm:NTDbyTD}
The logic $\DL(\partial^*)$ can be simulated by $\DL(\partial)$, and 
$\DL(\delta^*)$ can be simulated by $\DL(\delta)$, with respect to addition of facts.
\end{theorem}

The proof of this theorem is available
in the online appendix, pages 1--5.
It first shows, by induction on $n$, that, for any tagged literal $\alpha$ in $\DL(\partial^*)$ (or $\DL(\delta^*)$),
if $\alpha \in \T_{D+A} \uparrow n$ then $T(D)+A \vdash \alpha'$, 
where $\alpha'$ is the tagged literal in $\DL(\partial)$ (or $\DL(\delta)$) corresponding to $\alpha$.
Then, conversely, it shows that
if $\alpha' \in \T_{T(D)+A} \uparrow n$ then $D+A \vdash \alpha$.
Together, these establish that $\DL(\partial^*)$ is simulated by $\DL(\partial)$ under the transformation $T$ and,
similarly, that $\DL(\delta^*)$ is simulated by $\DL(\delta)$.

Thus $DL(\partial)$ is more (or equal) expressive than $DL(\partial^*)$
and
$DL(\delta)$ is more (or equal) expressive than $DL(\delta^*)$.

Notice that this result does not extend to simulation with respect to adding arbitrary defeasible theories
because, in that case, we can use the following defeasible theory as $A$ when $D = \emptyset$
and $p \in \Sigma$.

\begin{tabbing}
1234123412341234\=123412341234\=1234\=1234\kill

\>$r_1: \  \ \Rightarrow p$
\>\> $r_3: \  \ \Rightarrow \neg p$ \\

\>$r_2: \ \ \Rightarrow p$
\>\> $r_4: \ \ \Rightarrow \neg p$ \\

\>$r_1 > r_3$ \>\>  $r_2 > r_4$

\end{tabbing}
This theory distinguishes $\DL(\partial)$ from $\DL(\partial^*)$,
to demonstrate non-simulation in both directions,
and similarly for $\DL(\delta)$ and $\DL(\delta^*)$.

\section{Non-Team Defeat Simulates Team Defeat}

We define the transformation $T$ of $D$ to $T(D) = (F', R', >')$ as follows:
\begin{enumerate}
\item  \label{pt:facts}
The facts of $T(D)$ are the facts of $D$.
That is, $F' = F$.
\item  \label{pt:strict}
For each literal $q$, and each strict rule $r = (B \rightarrow  q)$ in $R$, $R'$ contains 
\[
\begin{array}{lrlrl}
ns(q): &   & \Rightarrow & \neg & strict(q) \\
s(r):    & B & \rightarrow &          & strict(q) \\
\end{array}
\]
and $ns(q) >' s(r)$.
\item  \label{pt:strict2}
For each literal $q$ defined by at least one strict rule in $R$,  $R'$ contains 
\[
\begin{array}{lrll}
& strict(q) & \rightarrow & q \\
\end{array}
\]

\item  \label{pt:dft}
For each ordered pair of opposing rules $r_i = (B_i \ARROW_{i} \non q)$ and $r_j = (B_j \ARROW_{j} q)$ in $R$,  
where 
$r_j$ is not a defeater, $R'$ contains
\[
\begin{array}{lrlrl}
R1_{ij}: & B_i &  \ARROW_i & \neg & d(r_i, r_j) \\
R2_{ij}: & B_j &  \Rightarrow && d(r_i, r_j) \\
R3_{ij}:  & strict(q) &  \Rightarrow && d(r_i, r_j) \\
          & d(r_i, r_j) & \Rightarrow && d(r_i) \\
        & fail(r_i) & \Rightarrow & & d(r_i) \\
NF_i:  & B_i & \Rightarrow & \neg & fail(r_i) \\
F_i:  &           & \Rightarrow && fail(r_i) \\
\end{array}
\]
and
$R2_{ij} >' R1_{ij}$ iff $r_j > r_i$, 
 $R3_{ij} >' R1_{ij}$ for every $i$ and $j$,
and $NF_i > F_i$ for every $i$. 

If there is no strict or defeasible rule $r_j$ for $q$ in $D$ then only the last three rules appear in $R'$, for each $i$.

\item  \label{pt:one}
For each literal $q$, and each strict or defeasible rule $r = (B \ARROW_r  q)$ in $R$, $R'$ contains 
\[
\begin{array}{lrll}
& B & \Rightarrow & one(q) \\
\end{array}
\]
\item  \label{pt:oneq}
For each literal $q$,  $R'$ contains 
\[
\begin{array}{lrll}
& one(q), d(r_1), \ldots, d(r_k) & \Rightarrow & q \\
\end{array}
\]
where $r_1, \ldots, r_k$ are the rules for $\non q$

\end{enumerate}

We say that a body $B$ \emph{fails} if $-d p$ is derived, for some $p \in B$,
and \emph{succeeds} if $+d B$ is derived,
where $d$ is the defeasible tag in the logic of interest.
We say a rule $r$ \emph{defeats} another $r'$ if they have opposing heads,
the body of $r$ succeeds and $r > r'$.

In the resulting theory $T(D)$,
$d(r_1, r_2)$ is derived iff
$r_1$ is defeated either because the body of $r_2$ succeeds and
$r_2 > r_1$, or because there is a strict opposing rule and its body is strictly provable.
$d(r)$ is derived iff some rule $r'$ defeats $r$, or the body of $r$ fails.
$one(q)$ is derived iff there is a strict or defeasible rule for $q$ and the body of that rule succeeds.
Thus, $q$ is derived if
there is a strict or defeasible rule for $q$ that succeeds and every rule for $\non q$ is defeated.
In this way, the transformed theory expresses team defeat.

Some elements of the definition deserve a more detailed explanation.
The first three points together define inference of $\pm\Delta$ from $T(D)$.
In point \ref{pt:strict} a defeasible rule is superior to a strict rule.
The effect of this somewhat counter-intuitive construction 
is to ensure $+\partial^* strict(q)$ is derived iff $+\Delta strict(q)$ is derived iff $+\Delta B$ is derived,
and $+\partial^* \neg strict(q)$ is derived iff $-\Delta B$ is derived.
It restricts the strict rule $s(r)$ to only be used for strict inferences, and not for defeasible inferences.
As a result, it ensures that all inferences to $q$ via point \ref{pt:strict2} are strict inferences.

Point \ref{pt:dft} identifies when a rule $r_i$ is defeated (as part of the process of inferring $q$).
$d(r_i, r_j)$ expresses that $r_i$ is defeated by $r_j$, and $d(r)$ expresses that $r$ is defeated.
If $r_i$ is strict and $+\Delta B_i$ is established then $r_i$ is not defeated (by $r_j$ or any other rule).
The use of $\ARROW_i$ in $R1_{ij}$ ensures this.
For $r_i$ to be defeated by $r_j$ we must have $+\partial^* B_j$.
If the stronger $+\Delta q$ can be established then $r_i$ is defeated unless,
by the above case, $r_i$ cannot be defeated.
This is expressed by $R3_{ij}$ with a defeasible rule so that, if the first case applies,
$+\Delta \neg d(r_i, r_j)$ is established and hence $+\partial^* d(r_i, r_j)$ cannot be derived.
However, $R3_{ij} >' R1_{ij}$ so that, in other circumstances, if $+\Delta q$ can be established then $r_i$ is defeated.
In the more normal case, if $+\partial^* B_j$ is established then $r_i$ is defeated by $r_j$
if either $-\partial^* B_i$ is established, or if $r_j > r_i$.
$R2_{ij}$ achieves this where the superiority relation in $T(D)$ has $R2_{ij} > R1_{ij}$, reflecting $r_j > r_i$.
Finally, the last three (classes of) rules of point \ref{pt:dft} identify that $r_i$ is defeated if its body $B_i$ fails.

Point \ref{pt:one} defines that $one(q)$ succeeds iff the body of some strict or defeasible rule for $q$ 
succeeds.
Point \ref{pt:oneq} then reflects the team defeat approach:
$q$ can be inferred if there is a strict or defeasible rule whose body succeeds ($one(q)$)
and every rule for $\non q$ is defeated ($d(r_1), \ldots, d(r_k)$).

\begin{example}
To see the operation of this transformation, we again consider the following theory $D$:
\begin{tabbing}
1234123412341234\=123412341234\=1234\=1234\kill

\>$r_1: \  \ \Rightarrow p$
\>\> $r_3: \  \ \Rightarrow \neg p$ \\

\>$r_2: \ \ \Rightarrow p$
\>\> $r_4: \ \ \Rightarrow \neg p$ \\

\>$r_1 > r_3$ \>\>  $r_2 > r_4$

\end{tabbing}
In $\DL(\partial)$ from $D$ we conclude $+\partial^* p$ and $-\partial^* \neg p$,
whereas
in $\DL(\partial^*)$ from $D$ we conclude $-\partial^* p$ and $-\partial^* \neg p$. 
$D$ does not contain any facts or strict rules, so parts \ref {pt:facts},  \ref {pt:strict}, and \ref {pt:strict2} 
do not contribute to $T(D)$.
$T(D)$ contains the following rules and superiority relation.
\smallskip
\begin{tabbing}
12\=3412341234123412341\=234123412341234123\=41234123412341234\=\kill

\>$R1_{13}:\  \Rightarrow \neg d(r_1, r_3)$
\>$R2_{13}:\  \Rightarrow  d(r_1, r_3)$
\>$R3_{13}:\  strict(p) \Rightarrow  d(r_1, r_3)$ \\

\>$R1_{14}:\  \Rightarrow \neg d(r_1, r_4)$
\>$R2_{14}:\  \Rightarrow  d(r_1, r_4)$
\>$R3_{14}:\  strict(p) \Rightarrow  d(r_1, r_4)$ \\

\>$R1_{23}:\  \Rightarrow \neg d(r_2, r_3)$
\>$R2_{23}:\  \Rightarrow  d(r_2, r_3)$
\>$R3_{23}:\  strict(p) \Rightarrow  d(r_2, r_3)$ \\

\>$R1_{24}:\  \Rightarrow \neg d(r_2, r_4)$
\>$R2_{24}:\  \Rightarrow  d(r_2, r_4)$
\>$R3_{24}:\  strict(p) \Rightarrow  d(r_2, r_4)$ \\

\>$R1_{31}:\  \Rightarrow \neg d(r_3, r_1)$
\>$R2_{31}:\  \Rightarrow  d(r_3, r_1)$
\>$R3_{31}:\  strict(p) \Rightarrow  d(r_3, r_1)$ \\

\>$R1_{32}:\  \Rightarrow \neg d(r_3, r_2)$
\>$R2_{32}:\  \Rightarrow  d(r_3, r_2)$
\>$R3_{32}:\  strict(p) \Rightarrow  d(r_3, r_2)$ \\

\>$R1_{41}:\  \Rightarrow \neg d(r_4, r_1)$
\>$R2_{41}:\  \Rightarrow  d(r_4, r_1)$
\>$R3_{41}:\  strict(p) \Rightarrow  d(r_4, r_1)$ \\

\>$R1_{42}:\  \Rightarrow \neg d(r_4, r_2)$
\>$R2_{42}:\  \Rightarrow  d(r_4, r_2)$
\>$R3_{42}:\  strict(p) \Rightarrow  d(r_4, r_2)$ \\
\ \\
\>$d(r_1, r_3) \Rightarrow d(r_1)$
\>$d(r_1, r_4) \Rightarrow d(r_1)$ 
\>$d(r_2, r_3) \Rightarrow d(r_2)$
\>$d(r_2, r_4) \Rightarrow d(r_2)$ \\
\ \\
\>$d(r_3, r_1) \Rightarrow d(r_3)$
\>$d(r_3, r_2) \Rightarrow d(r_3)$ 
\>$d(r_4, r_1) \Rightarrow d(r_4)$
\>$d(r_4, r_2) \Rightarrow d(r_4)$ \\
\ \\
\>$fail(r_1) \Rightarrow d(r_1)$
\>$fail(r_2) \Rightarrow d(r_2)$
\>$fail(r_3) \Rightarrow d(r_3)$
\>$fail(r_4) \Rightarrow d(r_4)$ \\

\>$NF_1:\  \Rightarrow \neg fail(r_1)$
\>$NF_2:\  \Rightarrow \neg fail(r_2)$
\>$NF_3:\  \Rightarrow \neg fail(r_3)$
\>$NF_4:\  \Rightarrow \neg fail(r_4)$ \\

\>$F_1:\  \Rightarrow fail(r_1)$
\>$F_2:\  \Rightarrow fail(r_2)$
\>$F_3:\  \Rightarrow fail(r_3)$
\>$F_4:\  \Rightarrow fail(r_4)$ \\

\ \\
\>$ \Rightarrow one(p)$
\>\>$ \Rightarrow one(\neg p)$ \\
\>$ \Rightarrow one(p)$
\>\>$ \Rightarrow one(\neg p)$ \\

\>$ one(p), d(r_3), d(r_4) \Rightarrow p$ 
\>\>$ one(\neg p), d(r_1), d(r_2) \Rightarrow \neg p$ \\
\ \\
\>$R2_{31} > R1_{31}$ 
\>\>$R3_{ij} > R1_{ij}$ for every opposing $i$ and $j$ \\
\>$R2_{42} > R1_{42}$ 
\>\>$NF_i > F_i$ for every $i$
\end{tabbing}
\smallskip

From $T(D)$ we can draw the conclusions $one(p)$, $+\partial^* d(r_3, r_1)$ and $+\partial^* d(r_4, r_2)$,
among others.
Consequently, we conclude $+\partial^* d(r_3)$ and $+\partial^* d(r_4)$,
and hence $+\partial^* p$.
This reflects the team defeat behaviour of $D$ within the non-team defeat logic $\DL(\partial^*)$.
\end{example}

In general, the size of $T(D)$ is quadratic in the size of $D$.
Thus it remains to establish that $T(D)$ simulates $D$ with respect to addition of facts.

\begin{theorem}   \label{thm:TDbyNTD}
The logic $\DL(\partial)$ can be simulated by $\DL(\partial^*)$,
and $\DL(\delta)$ can be simulated by $\DL(\delta^*)$, 
with respect to addition of facts.
\end{theorem}

The proof of this theorem is available
in the online appendix, from page 5.

As a result of Theorems \ref{thm:NTDbyTD} and \ref{thm:TDbyNTD},
the logics $\DL(\partial)$ and $\DL(\partial^*)$ have equal expressive power.
Similarly, $\DL(\delta)$ and $\DL(\delta^*)$ have equal expressive power.

\section{Ambiguity}

We now consider a different notion of expressiveness,
where simulation must be performed with respect to the addition of rules, not only facts.
We show that the ambiguity propagating logics cannot simulate the ambiguity blocking logics
with respect to additions of rules, and vice versa.
To show that a logic $L'$ cannot simulate $L$
it suffices to identify a theory $D$ and addition $A$ where
there is no $D'$ such that $D + A$ in $L$ and $D' + A$ in $L'$ have the same consequences.

\begin{theorem}   \label{thm:amb1}
Consider simulation with respect to addition of rules.
The logics $\DL(\partial)$ and $\DL(\partial^*)$
cannot be simulated by $\DL(\delta)$, nor by $\DL(\delta^*)$.
Conversely,
the logics $\DL(\delta)$ and $\DL(\delta^*)$ cannot be simulated by $\DL(\partial)$, nor by $\DL(\partial^*)$.
\end{theorem}

We first address the case of $\DL(\partial)$ and $\DL(\delta)$.
As mentioned above, it is sufficient to identify a single theory and addition that cannot be simulated.
Consider the theory $D$, with rules
\[
\begin{array}{lrll}
r_1: &  & \Rightarrow & p \\
r_2: &  & \Rightarrow & \neg p \\
\end{array}
\]
and consider an addition $A$ of rules
\[
\begin{array}{lrll}
r_3: &  & \Rightarrow & \neg p \\
r_4: &  & \Rightarrow & q \\
r_5: & \neg p & \Rightarrow & \neg q \\
\end{array}
\]

Then in $\DL(\partial)$ we have $D + A \vdash +\partial q$.
Suppose there is a theory $D'$ in $\DL(\delta)$ that simulates $D$ with respect to rules.
Then we must have $D' + A \vdash +\delta q$.
Furthermore, $-\Delta p, -\Delta \neg p, -\Delta q, -\Delta \neg q$
are consequences of $D + A$, and so are also consequences of $D' + A$ (and $D'$).
By the language separation condition, $D'$ does not contain any mention of $q$,
so the two rules $r_4$ and $r_5$ are the only rules under consideration for inferences about $q$.

Since $D' + A \vdash +\delta q$,
by the inference rule for $+\delta q$,
for (in this case) the rule $r_5$ for $\neg q$
either some literal in the body has no support (i.e. $-\supp \neg p$)
or the rule $r_4$ for $q$ over-rules $r_5$, that is, $r_4 > r_5$.
However, $r_4 > r_5$ is not part of $A$
and cannot be part of $D'$ (by the language separation condition).
Hence, we must have $D' + A \vdash -\supp \neg p$.

Now, by the inference rule for $-\supp$,
for every strict or defeasible rule for $\neg p$ either
some literal in the body has no support 
or there is a rule that can over-rule it.
Consider $r_3$.
No rule can over-rule it (by the language separation condition),
but the body of $r_3$ is empty.
This contradiction shows that no theory $D'$ in $\DL(\delta)$ simulates $D$ in $\DL(\partial)$.

$D + A$ has the same consequences, whether $\partial$ or $\partial^*$ is used.
Further, the argument is valid for $\delta^*$ as well as $\delta$.
Thus, neither $\partial$ nor $\partial^*$ can be simulated by either $\delta$ or $\delta^*$
with respect to the addition of rules.

The same theory and addition  can be used to show that the ambiguity blocking logics cannot simulate the ambiguity propagating logics.
Given $D$ and $A$ as above,
in $\DL(\delta)$ we have $D + A \vdash -\delta q$.
Suppose there is a theory $D'$ in $\DL(\partial)$ that simulates $D$ with respect to rules.
Then $D' + A \vdash -\partial q$.
As before, $-\Delta p, -\Delta \neg p, -\Delta q, -\Delta \neg q$ are consequences of $D' + A$ (and $D'$)
and, again, only rules $r_4$ (for $q$) and $r_5$ (for $\neg q$) directly affect inferences about $q$.
Furthermore, $D' + A \vdash -\partial \neg p$ since $-\delta \neg p$ is a consequence of $D+A$
and, hence, by the coherence property of $\DL$ (Proposition 2 of \cite{TOCL10}),
we cannot have $D' + A \vdash +\partial \neg p$.

Since $D' + A \vdash -\partial q$,
in the inference rule for $-\partial q$ only clause $-\partial.2.3$ can apply.
Thus we must have $D' + A \vdash +\partial \neg p$, by clause $-\partial.2.3.1$.
This gives us a contradiction, and hence no such $D'$ exists.
That is, $\DL(\delta)$ cannot be simulated by $\DL(\partial)$.

$D + A$ has the same consequences, whether $\delta$ or $\delta^*$ is used.
Further, the argument is valid for $\partial^*$ as well as $\partial$.
Thus, neither $\delta$ nor $\delta^*$ can be simulated by either $\partial$ or $\partial^*$
with respect to the addition of rules.

From these results, and the comments at the end of the section on simulating non-team defeat,
it is clear that simulation with respect to the addition of an arbitrary defeasible theory
is too strict to provide a viable notion of relative expressiveness.

Simulation with respect to addition of rules is stronger than simulation with respect to addition of facts,
(because addition of facts can be emulated by addition of strict rules with empty antecedents),
but is weaker than simulation with respect to full theories.
Thus the non-simulation results of this section do not necessarily extend to addition of facts.
That remains an open problem.

We could also consider simulation with respect to addition of rules, instead of facts, as the basis for a notion of relative expressiveness.
As we have seen, this notion is able to distinguish ambiguity propagating and blocking logics.
We would want to strengthen Theorems \ref{thm:NTDbyTD} and \ref{thm:TDbyNTD} to support this notion.

\section{Discussion}

The results of this paper are summarized in Figure \ref{fig:RE},
where an arrow from $d_1$ to $d_2$ expresses that $d_1$ can be polynomially simulated by $d_2$
with respect to the addition of facts.
Question marks between tags denote that the relationship is unknown.
This picture of relative expressiveness is quite different from the one for relative inference strength.
\begin{figure}   

\[
\begin{array}{rcl}
        \delta          &  \Longleftrightarrow  & \delta^*     \\
 \\
             ?             &                                 &    ?   \\
 \\
        \partial          &  \Longleftrightarrow  & \partial^*     \\
\end{array}
\]
\caption{Relative expressiveness of logics in $\DL$ using simulation wrt addition of facts}
\label{fig:RE}
\end{figure}

The relative inference strength of the logics in $\DL$ is described in Figure \ref{fig:RIS} (see \cite{TOCL10}).
$d_1 \subset d_2$ expresses that,
for any theory $D$,
the set of literals that are $+d_1$ consequences of $D$ is a subset of or equal to
the set of literals that are $+d_2$ consequences of $D$ and,
furthermore, there is a theory for which this containment is strict.
In addition,
the $-d_2$ consequences of $D$ are contained in the $-d_1$ consequences of $D$.

\begin{figure}   

\[
\begin{array}{rcccl}
 \Delta ~  \subset   ~    \delta^*          &  \subset  & \delta  ~  \subset  ~ \partial ~  \subset  ~ \supp &  \subset & \supp^* \\
 \\
                                                             &   \sesubset &                                                                               & \nesubset & \\
 \\
                                                            &                     &                                   \partial^*                        & \\
\end{array}
\]
\caption{Relative inference strength of logics in $\DL$}
\label{fig:RIS}
\end{figure}

It is interesting that $\partial$ and $\partial^*$ can simulate each other,
even though there is no relation between the two logics in terms of relative strength.
Furthermore, $\delta$ and $\delta^*$  can simulate each other even though,
in terms of relative inference strength,
$\delta^*$ is strictly weaker than $\delta$.
On the other hand, $\delta$ has weaker inference strength  than $\partial$ but yet
$\partial$ is unable to simulate $\delta$ under addition of rules and, 
similarly, $\delta^*$ has less inference strength than $\partial^*$ 
but $\partial^*$ is unable to simulate $\delta^*$ under addition of rules.
However $\delta$ is able to simulate the weaker in inference strength $\delta^*$.
Thus we see that relative expressiveness in defeasible logics is not directly related to
the relative inference strength of the logics.

This work is part of a long line of work addressing the relative expressibility of formalisms,
of which we will mention just a few.
Interpretation of one theory by another in classical logic (for example, \cite{Shoenfield})
essentially maps functions in one language into terms from another
in such a way that the axioms of one theory map to theorems in the other.
This extends easily to the interpretation of theories in different, but similar, logics.
This technique provides a basis for transferring results on consistency and decidability
from one theory to another.
The idea was used in \cite{CoxMT92} to transfer complexity results for CLP languages.
Similarly, the idea of a conservative extension and extension by definitions of a theory  \cite{Shoenfield}
can be used to establish that some programming language features do not extend the expressive power
of a language \cite{Landin,Felleisen}.
In general, any sequential programming language can simulate another (the ``Turing tarpit'')
but, by requiring that the mapping of one language into another be homomorphic 
(which enforces a preservation of structure) 
and observing the behaviour in any context,
a meaningful notion of relative expressiveness can be developed \cite{Felleisen}.
These ideas were extended for concurrent languages \cite{Shapiro89,dBP1,dBP2} where, in addition,
it was required that parallel composition and nondeterministic choice in the simulated language
were represented by parallel composition and nondeterministic choice in the simulating language.
A more general treatment is \cite{Shapiro91}.
More recently, \cite{Janhunen06} investigated relative expressiveness
for logic programs using a polynomial bound on the translation and a weak form of modularity.

There has also been some related work in defeasible logic.
Early work \cite{TOCL01}  on $\DL(\partial)$ 
demonstrated that some features of the logic --
facts, defeaters and the superiority relation --
do not add to the expressiveness to that logic\footnote{
These results do not all extend to ambiguity propagating logics \cite{Lam.2011}.
}.
Furthermore,
the idea of simulation with respect to additions
is similar to the idea of modular transformation in \cite{TOCL01}.
In \cite{MG99}, failure operators were added to $\DL(\partial)$
and shown to be a conservative extension.
In \cite{MG99,TPLP06}, a simulation of $\DL(\partial)$ in logic programs under the Kunen semantics was shown,
and in \cite{TPLP06} it was shown that this transformation 
does not provide a simulation by logic programs under the stable model semantics.
 
\section{Conclusion}

We have introduced a notion of relative expressiveness for defeasible logics,
based on simulation with respect to addition of facts,
and shown that it is not too strict.
The simulation of a logic $\DL(\partial)$ with team defeat by a logic  $\DL(\partial^*)$
without team defeat is a surprising demonstration of that fact.
However, it remains an open question whether there is a relative expressiveness relationship
between the ambiguity blocking and propagating logics.

We have also investigated alternative notions of relative expressiveness,
and seen that simulation with respect to rules is not too lenient.
It remains to determine whether it is too strict or not.
We have already seen, in the section on ambiguity, that simulation with respect to full defeasible theories is too strict.

{\bf Acknowledgements:}  The author thanks the referees for their careful reviewing.

%\newpage
\bibliographystyle{acmtrans}
\bibliography{relative_expressiveness}

\end{document}